\documentclass[10pt, a4paper]{article}
\usepackage{lrec2016}
\usepackage{graphicx}

\usepackage{epstopdf}
\usepackage[latin1]{inputenc}
\usepackage{url}
 
 \usepackage{lingmacros} 
 \usepackage{mdframed}
 \usepackage{tabularx}
 \usepackage{subcaption}
 \usepackage{textcomp}
 \usepackage{comment}
  
\graphicspath{ {./figures} }
\usepackage{rotating}
\usepackage{array}  

\usepackage{multirow}

\usepackage{tcolorbox}

\newcolumntype{P}[1]{>{\centering\arraybackslash}p{#1}} 

\title{InScript: Narrative texts annotated with script information}

\name{Ashutosh Modi, Tatjana Anikina, Simon Ostermann, Manfred Pinkal}

\address{ Universit{\"a}t des Saarlandes \\
               Saarland, 66123, Germany \\
               \{ashutosh, tatianak, simono, pinkal\}@coli.uni-saarland.de\\}

\abstract{
This paper presents the InScript corpus (Narrative Texts \textbf{In}stantiating \textbf{Script} structure). InScript is a corpus of 1,000 stories centered around 10 different scenarios. Verbs and noun phrases are annotated with event and participant types, respectively. Additionally, the text is annotated with coreference information. The corpus shows rich lexical variation and will serve as a unique resource for the study of the role of script knowledge in natural language processing.  
 \\ 
\newline 
 }

\begin{document}

\maketitleabstract

\section{Motivation}

A \textit{script} is ``a standardized sequence of events that
describes some stereotypical human activity such
as going to a restaurant or visiting a doctor'' \cite{Barr1981}. 
Script events describe an action/activity along with the involved participants.   
For example, in the script describing \textsc{a visit to a restaurant}, typical events are \textsc{entering the restaurant}, \textsc{ordering food} or \textsc{eating}. Participants in this scenario can include animate objects like the \textsc{waiter} and the \textsc{customer}, as well as inanimate objects such as \textsc{cutlery} or \textsc{food}.

Script knowledge has been shown to play an important role in text understanding (\newcite{cullingford1978script}, \newcite{miikkulainen1995script}, \newcite{mueller2004understanding}, \newcite{Chambers2008}, \newcite{Chambers2009}, \newcite{modi2014inducing}, \newcite{rudinger2015learning}). It guides the expectation of the reader, supports coreference resolution as well as common-sense knowledge inference and enables the appropriate embedding of the current sentence into the larger context. Figure \ref{fig:story} shows the first few sentences of a story describing the scenario \textsc{taking a bath}. Once the \textsc{taking a bath} scenario is evoked by the noun phrase (NP)  ``a bath'', the reader can effortlessly interpret the definite NP ``the faucet'' as an implicitly present standard participant of the \textsc{taking a bath} script.  Although in this story, ``entering the bath room'', ``turning on the water'' and ``filling the tub'' are explicitly mentioned, a reader could nevertheless have inferred the ``turning on the water'' event, even if it was not explicitly mentioned in the text. Table \ref{table:template} gives an example of typical events and participants for the script describing the scenario \textsc{taking a bath}.

\begin{figure}[h]
\begin{tcolorbox}
\begin{footnotesize}
I was sitting on my couch when I decided that I hadn't taken a bath in a while so I stood up and walked to the bathroom where I turned on the faucet in the sink and began filling the bath with hot water. \\

While the tub was filling with hot water I put some bubble bath into the stream of hot water coming out of the faucet so that the tub filled with not only hot water[...]
\end{footnotesize}
\end{tcolorbox}
\caption{An excerpt from a story on the \textsc{taking a bath} script.}
\label{fig:story}
\end{figure}

A systematic study of the influence of script knowledge in texts is far from trivial. Typically, text documents (e.g.\ narrative texts)  describing various scenarios evoke many different scripts, making it difficult to study the effect of a single script. Efforts have been made to collect scenario-specific script knowledge via crowdsourcing, for example the OMICS and SMILE corpora (\newcite{singh2002open}, \newcite{Regneri:2010}, \newcite{Regneri2013}), but these corpora describe script events in a pointwise telegram style rather than in full texts. 

This paper presents the InScript \footnote{\label{corpusLink}The corpus can be downloaded at: \url{http://www.sfb1102.uni-saarland.de/?page_id=2582} } corpus (Narrative Texts \textbf{In}stantiating \textbf{Script} structure). It is a corpus of simple narrative texts in the form of stories, wherein each story is centered around a specific scenario. The stories have been collected via Amazon Mechanical Turk (M-Turk)\footnote{\label{mturkLink} \url{https://www.mturk.com}}. In this experiment, turkers were asked to write down a concrete experience about a bus ride, a grocery shopping event etc. We concentrated on 10 scenarios and collected 100 stories per scenario, giving a total of 1,000 stories with about 200,000 words. Relevant verbs and noun phrases in all stories are annotated with \textit{event types} and \textit{participant types} respectively. Additionally, the texts have been annotated with coreference information in order to facilitate the study of the interdependence between script structure and coreference.

\begin{figure}[h]
\centering
\includegraphics[width=\columnwidth]{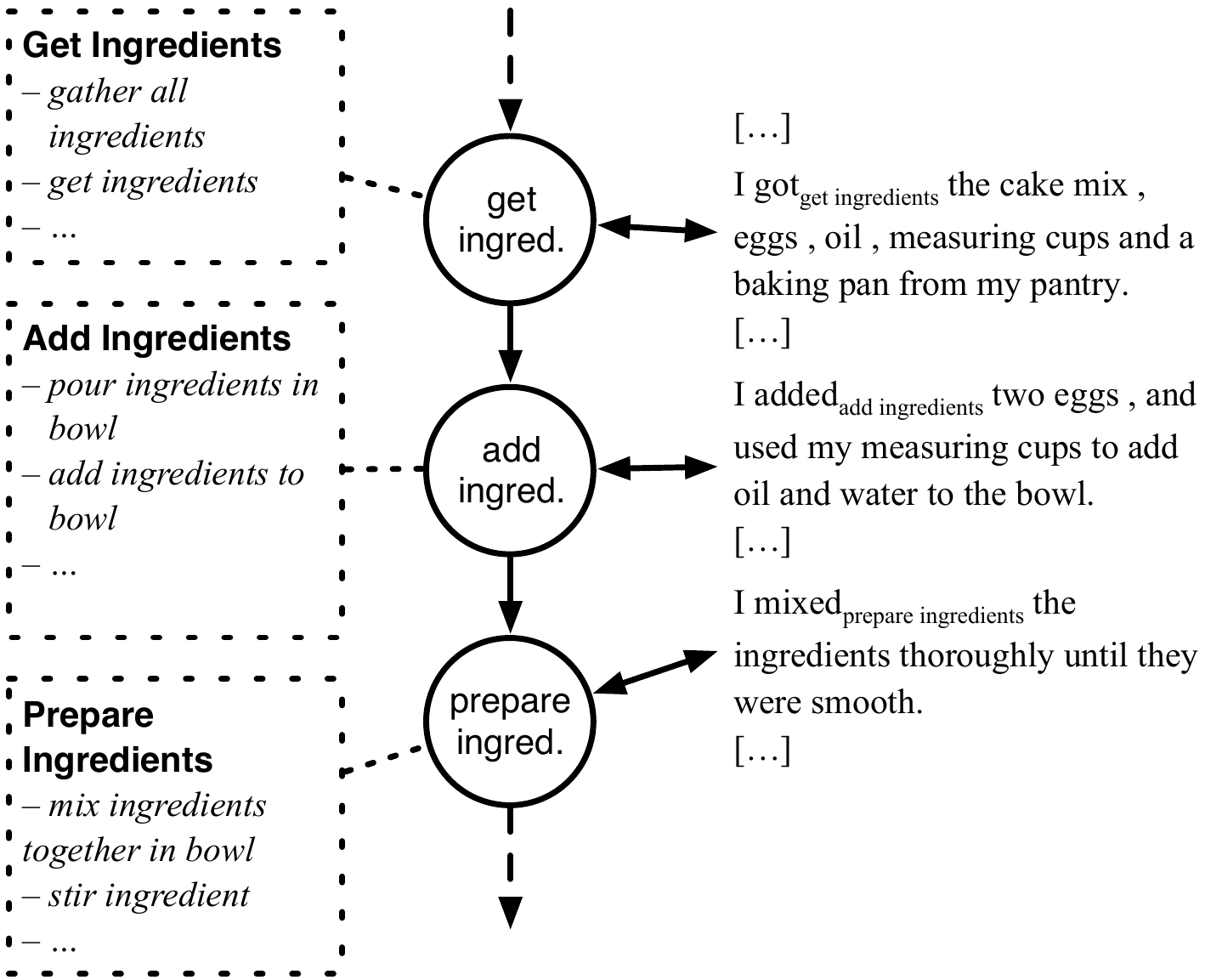}
\caption{Connecting DeScript and InScript: an example from the \textsc{Baking a cake} scenario (InScript participant annotation is omitted for better readability).}
\label{fig:scr_ex}
\end{figure}

The InScript corpus is a unique resource that provides a basis for studying various aspects of the role of script knowledge in language processing by humans. The acquisition of this corpus is part of a larger research effort that aims at using script knowledge to model the surprisal and information density in written text. Besides InScript, this project also released a corpus of generic descriptions of script activities called DeScript (for \textbf{De}scribing \textbf{Script} Structure, \newcite{Wanzare2016}). DeScript contains a range of short and textually simple phrases that describe script events in the style of OMICS or SMILE (\newcite{singh2002open}, \newcite{Regneri:2010}). These generic telegram-style descriptions are called \textit{Event Descriptions} (EDs); a sequence of such descriptions that cover a complete script is called an \textit{Event Sequence Description} (ESD). Figure \ref{fig:scr_ex} shows an excerpt of a script in the \textsc{baking a cake} scenario. The figure shows event descriptions for 3 different events in the DeScript corpus (left) and fragments of a story in the InScript corpus (right) that instantiate the same event type.


\begin{table}[h]
\centering
\small
\begin{tabular}{|l|l|}                         
\hline
Event types & Participant types                \\
\hline
\textsc{ScrEv\_take\_clean}& \textsc{\textbf{ScrPart\_bath}}   \\
\textsc{\_clothes} & \\
\textsc{ScrEv\_prepare\_bath}  & \textsc{ScrPart\_bath\_means     }      \\
\textsc{ScrEv\_enter\_bathroom}  & \textsc{ScrPatr\_bather      }       \\
\textsc{ScrEv\_turn\_water\_on } & \textsc{ScrPart\_bathroom     }      \\
\textsc{ScrEv\_check\_temp}  & \textsc{ScrPart\_bathtub    }                \\
(temperature) & \\
\textsc{ScrEv\_close\_drain}  & \textsc{ScrPart\_body\_part  }                 \\
\textsc{ScrEv\_wait}  & \textsc{ScrPart\_clothes}  \\
\textsc{ScrEv\_turn\_water\_off}  & \textsc{ScrPart\_drain }                     \\
\textsc{ScrEv\_put\_bubble}& \textsc{ScrPart\_hair  }         \\
\textsc{\_bath\_scent}  & \\
\textsc{ScrEv\_undress}  & \textsc{ScrPart\_hamper }           \\
\textsc{ScrEv\_sink\_water}  & \textsc{ScrPart\_in-bath}\\&\textsc{\_entertainment } \\&(candles, music, books)     \\  
\textsc{ScrEv\_relax}  & \textsc{\textbf{ScrPart\_plug} }      \\ 
\textsc{ScrEv\_apply\_soap}  & \textsc{\textbf{ScrPart\_shower}}  \\&(as bath equipment)      \\
\textsc{ScrEv\_wash}  & \textsc{ScrPart\_tap (knob) }      \\  
\textsc{ScrEv\_open\_drain}  & \textsc{ScrPart\_temperature }      \\
\textsc{ScrEv\_get\_out\_bath}  & \textsc{ScrPart\_towel }      \\  
\textsc{\textbf{ScrEv\_get\_towel}}  & \textsc{ScrPart\_washing\_tools}  \\& (washcloth, soap)      \\ 
\textsc{ScrEv\_dry}  & \textsc{ScrPart\_water}       \\ 
\textsc{ScrEv\_put\_after\_shower}  & \\ 
\textsc{ScrEv\_get\_dressed}  &      \\
\textsc{ScrEv\_leave}  &      \\  
\textsc{ScrEv\_air\_bathroom}  &       \\ 
\hline           
\end{tabular}
\caption{Bath scenario template (labels added in the second phase of annotation are marked in bold).}
\label{table:template}
\end{table}

\section{Data Collection}
\subsection{Collection via Amazon M-Turk}
We selected 10 scenarios from different available scenario lists (e.g.\ \newcite{Regneri:2010} , \newcite{VanDerMeer2009}, and the OMICS corpus \cite{singh2002open}), including scripts of different complexity (\textsc{Taking a bath} vs. \textsc{Flying in an airplane}) and specificity (\textsc{Riding a public bus} vs. \textsc{Repairing a flat bicycle tire}). For the full scenario list see Table \ref{table:scenarios}. 

\begin{table*}[t]
\centering
\small
\resizebox{2\columnwidth}{!}{
\begin{tabular}{|P{6cm}|P{1cm}|P{2cm}|P{2cm}|P{2cm}|P{2cm}|}
\hline
Scenario Name & \#Stories & Avg. Sentences Per Story & Avg. Word Type Per Story & Avg. Word Count Per Story & Avg. Word Type Overlap\\
\hline
\textsc{Riding in a public bus (Bus)} &  92 & 12.3 (4.1)  & 97.4 (23.3)  & 215.1 (69.7) & 35.7 (7.5) \\
\hline
\textsc{Baking a cake (Cake)} &  97	& 13.6 (4.7)  & 102.7 (23.7)  & 235.5 (78.5) & 39.5 (8.1) \\
\hline
\textsc{Taking a bath (Bath)} & 94 & 11.5 (2.6)  & 91.9 (13.1)  & \textbf{\textit{197.5 (34.5)}}  & 37.9 (6.3)\\
\hline
\textsc{Going grocery shopping (Grocery)}& 95 & 13.1 (3.7)  & 102.9 (19.9)  & 228.3 (58.8) & 38.6 (7.8) \\
\hline
\textsc{Flying in an airplane (Flight)}&  86 & \textbf{14.1 (5.6)}  & \textbf{113.6 (30.9)}  & \textbf{251.2 (99.1)} & \textbf{40.9 (10.3)} \\
\hline
\textsc{Getting a haircut (Haircut)}& 88 & 13.3 (4.0)  & 100.6 (19.3)  & 227.2 (63.4) & 39.0 (7.9) \\
\hline
\textsc{Borrowing a book from the library (Library)}&  93 & 11.2 (2.5) 	 & \textbf{\textit{88.0 (14.1)}} 	& 200.7 (43.5) & 34.9 (5.5)	\\
\hline
\textsc{Going on a train (Train)}&  87 & 12.3 (3.4)  & 96.3 (19.2)  & 210.3 (57.0) & 35.3 (6.9) \\
\hline
\textsc{Repairing a flat bicycle tire (Bicycle)}&  87 & 11.4 (3.6)  & 88.9 (15.0)  & 203.0 (53.3) & 33.8 (5.2) \\
\hline
\textsc{Planting a tree (Tree)} &  91	& \textbf{\textit{11.0 (3.6)}}  & 93.3 (19.2)  & 201.5 (60.3) & \textbf{\textit{34.0 (6.6)}} \\
\hline
\textit{Average} &  \textit{91} & \textit{12.4} & \textit{97.6} & \textit{216.9} & \textit{37.0} \\
\hline
\end{tabular}
}
\caption{Corpus statistics for different scenarios (standard deviation given in parentheses). The maximum per column is highlighted in \textbf{boldface}, the minimum in \textbf{\textit{boldface italics}}.}
\label{table:scenarios}
\end{table*}%

Texts were collected via the Amazon Mechanical Turk platform, which provides an opportunity to present an online task to humans (a.k.a. \textit{turkers}).
In order to gauge the effect of different M-Turk instructions on our task, we first conducted pilot experiments with different variants of instructions explaining the task. 
We finalized the instructions for the full data collection, asking the turkers to describe a scenario in form of a story \textit{as if explaining it to a child} and to use a minimum of 150 words. The selected instruction variant resulted in comparably simple and explicit scenario-related stories. In the future we plan to collect more complex stories using different instructions.
In total 190 turkers participated. All turkers were living in the USA and native speakers of English. We paid USD \$0.50\ per story to each turker. On average, the turkers took 9.37 minutes per story with a maximum duration of 17.38 minutes. 

\subsection{Data Statistics}
Statistics for the corpus are given in Table \ref{table:scenarios}. On average, each story has a length of 12 sentences and 217 words with 98 word types on average. Stories are coherent and concentrate mainly on the corresponding scenario. Neglecting auxiliaries, modals and copulas, on average each story has 32 verbs, out of which 58\% denote events related to the respective scenario. As can be seen in Table \ref{table:scenarios}, there is some variation in stories across scenarios: The \textsc{flying in an airplane} scenario, for example, is most complex in terms of the number of sentences, tokens and word types that are used. This is probably due to the inherent complexity of the scenario: Taking a flight, for example, is more complicated and takes more steps than taking a bath. The average count of sentences, tokens and types is also very high for the \textsc{baking a cake} scenario. Stories from the scenario often resemble cake recipes, which usually contain very detailed steps, so people tend to give more detailed descriptions in the stories. 

For both \textsc{flying in an airplane} and \textsc{baking a cake}, the standard deviation is higher in comparison to other scenarios. This indicates that different turkers described the scenario with a varying degree of detail and can also be seen as an indicator for the complexity of both scenarios. In general, different people tend to describe situations subjectively, with a varying degree of detail. 

In contrast, texts from the \textsc{taking a bath} and \textsc{planting a tree} scenarios contain a relatively smaller number of sentences and fewer word types and tokens. Both planting a tree and taking a bath are simpler activities, which results in generally less complex texts.

The average pairwise word type overlap can be seen as a measure of lexical variety among stories: If it is high, the stories resemble each other more. We can see that stories in the \textsc{flying in an airplane} and \textsc{baking a cake} scenarios have the highest values here, indicating that most turkers used a similar vocabulary in their stories.

In general, the response quality was good. We had to discard 9\% of the stories as these lacked the quality we were expecting. In total, we selected 910 stories for annotation. 

\section{Annotation}
This section deals with the annotation of the data. We first describe the final annotation schema. Then, we describe the iterative process of corpus annotation and the refinement of the schema. This refinement was necessary due to the complexity of the annotation.

\subsection{Annotation Schema}
\label{sec:schema}
For each of the scenarios, we designed a specific annotation template. A \textit{script template} consists of scenario-specific event and participant labels. An example of a template is shown in Table \ref{table:template}. All NP heads in the corpus were annotated with a participant label; all verbs were annotated with an event label. 
For both participants and events, we also offered the label \textsc{unclear} if the annotator could not assign another label. We additionally annotated coreference chains between NPs. Thus, the process resulted in three layers of annotation: event types, participant types and coreference annotation. These are described in detail below. 

\paragraph{Event Type}\ \\
As a first layer, we annotated event types. There are two kinds of event type labels, scenario-specific event type labels and general labels. The general labels are used across every scenario and mark general features, for example whether an event belongs to the scenario at all. For the scenario-specific labels, we designed an unique template for every scenario, with a list of script-relevant event types that were used as labels. Such labels include for example \textsc{ScrEv\_close\_drain} in \textsc{taking a bath} as in Example \ref{ex:label} (see Figure \ref{table:template} for a complete list for the \textsc{taking a bath} scenario)

\enumsentence{I start by \textit{closing}$_{\textsc{\scriptsize ScrEv\_close\_drain}}$ the drain at the bottom of the tub.}
\label{ex:label}

The general labels that were used in addition to the script-specific labels in every scenario are listed below:

\begin{itemize}
\item\textsc{ScrEv\_other}. An event that belongs to the scenario, but its event type occurs too infrequently (for details, see below, Section \ref{sec:modifications}). We used the label  ``other" because event classification would become too finegrained otherwise. \\ \textit{Example:} After I am dried I put my new clothes on and \textit{clean up}$_{\textsc{\scriptsize ScrEv\_other}}$ the bathroom.
\item\textsc{RelNScrEv}. Related non-script event. An event that \textit{can} plausibly happen during the execution of the script and is related to it, but that is not part of the script. \\ \textit{Example:} After finding on what I wanted to wear, I went into the bathroom and  \textit{shut}$_{\textsc{\scriptsize RelNScrEv}}$ the door.
\item\textsc{UnrelEv}. An event that is unrelated to the script. \\ \textit{Example:} I sank into the bubbles and \textit{took}$_{\textsc{\scriptsize UnrelEv}}$ a deep breath.
\end{itemize}

Additionally, the annotators were asked to annotate verbs and phrases that evoke the script without explicitly referring to a script event with the label \textsc{Evoking}, as shown in Example \ref{ex:evoke}. 
\enumsentence{Today I \textit{took a bath}$_{\textsc{\scriptsize Evoking}}$ in my new apartment.}
\label{ex:evoke}

\begin{figure*}
\centering
\fbox{\includegraphics[trim=0cm 15cm 0cm 0cm,width=\textwidth, scale=.5]{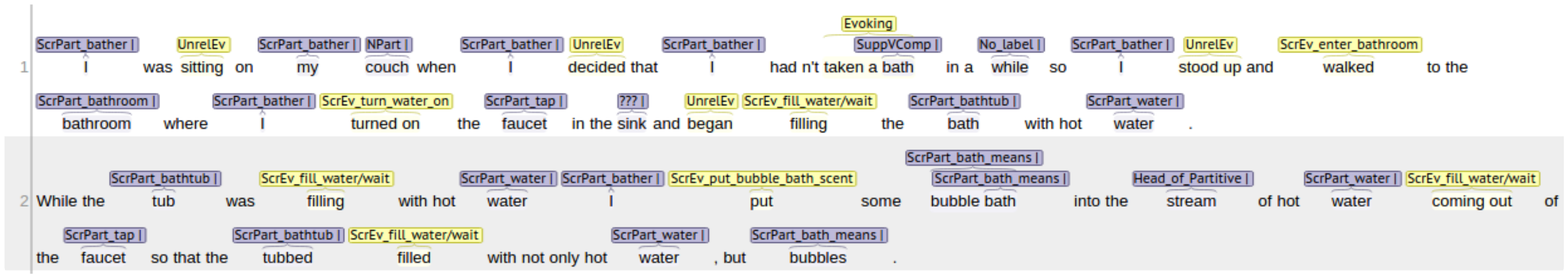}}
\caption[BathAnnotation]{Sample event and participant annotation for the \textsc{taking a bath} script.}
\label{fig:annotation_example}
\end{figure*}

\paragraph{Participant Type}\ \\
As in the case of the event type labels, there are two kinds of participant labels: general labels and scenario-specific labels. The latter are part of the scenario-specific templates, e.g.\ \textsc{ScrPart\_drain} in the \textsc{taking a bath} scenario, as can be seen in Example \ref{ex:partlabel}.

\enumsentence{I start by closing the \textit{drain}$_{\textsc{\scriptsize ScrPart\_drain}}$ at the bottom of the tub.}
\label{ex:partlabel}

The general labels that are used across all scenarios mark noun phrases with scenario-independent features. There are the following general labels:

\begin{itemize}
\item\textsc{ScrPart\_other}. A participant that belongs to the scenario, but its participant type occurs only infrequently. \\ \textit{Example:}  I find my \textit{bath mat}$_{\textsc{\scriptsize ScrPart\_other}}$ and lay it on the floor to keep the floor dry.

\item \textsc{NPart}. Non-participant. A referential NP that does not belong to the scenario. \\ \textit{Example:}  I washed myself carefully because I did not want to spill water onto the \textit{floor}$_{\textsc{\scriptsize NPart}}$.labeled

\item \textsc{SuppVComp}. A support verb complement. For further discussion of this label, see Section \ref{sec:specialcases} \\ \textit{Example:}  I sank into the bubbles and took a deep \textit{breath}$_{\textsc{\scriptsize SuppVComp}}$.

\item \textsc{Head\_of\_Partitive}. The head of a partitive or a partitive-like construction. For a further discussion of this label cf.\ Section \ref{sec:specialcases} \\ \textit{Example:}  I grabbed a \textit{bar}$_{\textsc{\scriptsize Head\_of\_Partitive}}$ of soap and lathered my body.

\item\textsc{No\_label}. A non-referential noun phrase that cannot be labeled with another label.\ \textit{Example:}  I sat for a \textit{moment}$_{\textsc{\scriptsize No\_label}}$, relaxing, allowing the warm water to sooth my skin.\\
\end{itemize}

All NPs labeled with one of the labels \textsc{SuppVComp}, \textsc{Head\_of\_Partitive} or \textsc{No\_label} are considered to be non-referential. \textsc{No\_label} is used mainly in four cases in our data: non-referential time expressions (\textit{in a \textbf{while}}, \textit{a million \textbf{times} better}), idioms (\textit{no \textbf{matter} what}), the non-referential ``it'' (\textit{\textbf{it} felt amazing}, \textit{\textbf{it} is better}) and other abstracta (\textit{a \textbf{lot} better}, \textit{a little \textbf{bit}}).

In the first annotation phase, annotators were asked to mark verbs and noun phrases that have an event or participant type, that is \textit{not} listed in the template, as \textsc{MissScrEv}/ \textsc{MissScrPart} (missing script event or participant, resp.). These annotations were used as a basis for extending the templates (see Section \ref{sec:modifications}) and replaced later by newly introduced labels or \textsc{ScrEv\_other} and \textsc{ScrPart\_other} respectively.

\paragraph{Coreference Annotations}\ \\
All noun phrases were annotated with coreference information indicating which entities denote the same discourse referent. The annotation was done by linking heads of NPs (see Example \ref{ex:coref}, where the links are indicated by coindexing). As a rule, we assume that each element of a coreference chain is marked with the same participant type label.

\enumsentence  {\textit{I}$ _{\textsc{\scriptsize Coref1}}$ washed \textit{my}$ _{\textsc{\scriptsize Coref1}}$ entire \textit{body}$ _{\textsc{\scriptsize Coref2}}$, starting with \textit{my}$ _{\textsc{\scriptsize Coref1}}$ \textit{face}$ _{\textsc{\scriptsize Coref3}} $ and ending with the \textit{toes}$ _{\textsc{\scriptsize Coref4}} $. \textit{I}$ _{\textsc{\scriptsize Coref1}}$ always wash \textit{my}$ _{\textsc{\scriptsize Coref1}}$ \textit{toes}$_{\textsc{\scriptsize Coref4}}$ very thoroughly ... 
}  
\label{ex:coref}

The assignment of an entity to a referent is not always trivial, as is shown in Example \ref{ex:coref_groups}. There are some cases in which two discourse referents are grouped in a plural NP. In the example, \textit{those things} refers to the group made up of \textit{shampoo}, \textit{soap} and \textit{sponge}. In this case, we asked annotators to introduce a new coreference label, the name of which indicates which referents are grouped together (\textsc{Coref\_group\_washing\_tools}). All NPs are then connected to the group phrase, resulting in an additional coreference chain.


\enumsentence  {\textit{I}$ _{\textsc{\scriptsize Coref1}}$ made sure that \textit{I}$ _{\textsc{\scriptsize Coref1}}$ have \textit{my}$ _{\textsc{\scriptsize Coref1}}$ \textit{shampoo}$ _{\textsc{\scriptsize Coref2 + Coref\_group\_washing\_tools}}$, \textit{soap}$_{\textsc{\scriptsize Coref3 + Coref\_group\_washing\_tools}}$ and \textit{sponge}$ _{\textsc{\scriptsize Coref4 + Coref\_group\_washing\_tools}}$ ready to get in. Once \textit{I}$ _{\textsc{\scriptsize Coref1}}$ have those \textit{things}$ _{\textsc{\scriptsize Coref\_group\_washing\_tools}}$ \textit{I}$ _{\textsc{\scriptsize Coref1}}$ sink into the bath. ... \textit{I}$ _{\textsc{\scriptsize Coref1}}$ applied some \textit{soap}$ _{\textsc{\scriptsize Coref3}}$ on \textit{my}$ _{\textsc{\scriptsize Coref1}}$ body and used the \textit{sponge}$ _{\textsc{\scriptsize Coref4}}$ to scrub a bit. ... \textit{I}$ _{\textsc{\scriptsize Coref1}}$ rinsed the \textit{shampoo}$ _{\textsc{\scriptsize Coref2}}$. 
\label{ex:coref_groups}
}
Example \ref{ex:coref_groups} thus contains the following coreference chains:
\enumsentence{
{\textsc{\scriptsize Coref1}}: I $\rightarrow$ I $\rightarrow$ my $\rightarrow$ I $\rightarrow$ I $\rightarrow$ I $\rightarrow$ my $\rightarrow$ I \\
{\textsc{\scriptsize Coref2}}: shampoo $\rightarrow$ shampoo \\
{\textsc{\scriptsize Coref3}}: soap $\rightarrow$ soap \\
{\textsc{\scriptsize Coref4}}: sponge $\rightarrow$ sponge \\
\textsc{\scriptsize{Coref\_group\_washing\_ tools}}: shampoo $\rightarrow$ soap $\rightarrow$ sponge $\rightarrow$ things 
}

\subsection{Development of the Schema}
The templates were carefully designed in an iterated process. For each scenario, one of the authors of this paper provided a preliminary version of the template based on the inspection of some of the stories. For a subset of the scenarios, preliminary templates developed at our department for a psycholinguistic experiment on script knowledge were used as a starting point. Subsequently, the authors manually annotated 5 randomly selected texts for each of the scenarios based on the preliminary template. Necessary extensions and changes in the templates were discussed and agreed upon. Most of the cases of disagreement were related to the granularity of the event and participant types. We agreed on the \textit{script-specific functional equivalence} as a guiding principle. For example, reading a book, listening to music and having a conversation are subsumed under the same event label in the \textsc{flight} scenario, because they have the common function of in-flight entertainment in the scenario. In contrast, we assumed different labels for the cake tin and other utensils (bowls etc.), since they have different functions in the \textsc{baking a cake} scenario and accordingly occur with different script events. 

Note that scripts and templates as such are not meant to describe an activity as exhaustively as possible and to mention all steps that are logically necessary. Instead, scripts describe cognitively prominent events in an activity. An example can be found in the \textsc{flight} scenario. While more than a third of the turkers mentioned the event of fastening the seat belts in the plane (\textsc{buckle\_seat\_belt}), no person wrote about \textit{undoing} their seat belts again, although in reality both events appear equally often. Consequently, we added an event type label for buckling up, but no label for undoing the seat belts.



\subsection{First Annotation Phase}
We used the WebAnno annotation tool \cite{YimamGCB13} for our project. The stories from each scenario were distributed among four different annotators. In a calibration phase, annotators were presented with some sample texts 
 for test annotations; the results were discussed with the authors. Throughout the whole annotation phase, annotators could discuss any emerging issues with the authors. All annotations were done by undergraduate students of computational linguistics. The annotation was rather time-consuming due to the complexity of the task, and thus we decided for single annotation mode. To assess annotation quality, a small sample of texts was annotated by all four annotators and their inter-annotator agreement was measured (see Section \ref{sec:iaa}). It was found to be sufficiently high.

Annotation of the corpus together with some pre- and post-processing of the data required about 500 hours of work. All stories were annotated with event and participant types (a total of 12,188 and 43,946 instances, respectively). On average there were 7 coreference chains per story with an average length of 6 tokens.

\begin{figure*}[htp]

\begin{subfigure}[b]{0.6\linewidth}
\centering
\small
\begin{tabular}{|c|c|c|c|c|}
\hline
&\multicolumn{4}{c|}{Average Fleiss' Kappa}
\\
\hline
       & \multicolumn{2}{c|}{All Labels} & \multicolumn{2}{c|}{Script Labels} \\ \hline
 Scenario                 & Events                 & Participants          & Events                  &  
                 Participants                      \\ \hline
\textsc{Bus}              & 0.68                   & 0.74                  & 0.76                    & 0.74                    \\ \hline
\textsc{Cake}             & 0.61                   & 0.76                  & 0.64                    & 0.75                    \\ \hline
\textsc{Flight}          & 0.65                   & 0.70                  & 0.62                    & 0.69                    \\ \hline
\textsc{Grocery}          & 0.64                   & 0.80                  & 0.73                    & 0.80                    \\ \hline
\textsc{Haircut}          & 0.64                   & 0.84                  & 0.67                    & 0.86                    \\ \hline
\textsc{Tree}             & 0.59                   & 0.76                  & 0.63                    & 0.76                    \\ \hline
\textit{Average} & \textit{0.64}          & \textit{0.77}         & \textit{0.68}           & \textit{0.77}           \\ \hline
\end{tabular}
\caption{Average Fleiss' Kappa.}
\label{table:fleiss_kappa}
\end{subfigure} %
%
%
\begin{subfigure}[b]{0.36\linewidth}
\centering
\small
\begin{tabular}{|c|c|}
\hline
Scenario         & \%Coreference Agreement \\
\hline
\textsc{Bus}              & 88.9                    \\
\hline
\textsc{Cake}             & 94.7                    \\
\hline
\textsc{Flight}          & 93.6                    \\
\hline
\textsc{Grocery}          & 93.4                    \\
\hline
\textsc{Haircut}          & 94.3                    \\
\hline
\textsc{Tree}             & 78.3                    \\
\hline
\textit{Average} & \textit{90.5} \\ 
\hline        
\end{tabular}
\caption{Coreference agreement.}
\label{table:coref_stat}
\end{subfigure}
\caption{Inter-annotator agreement statistics.}
\label{table:interanno}
\end{figure*}

\subsection{Modification of the Schema}
\label{sec:modifications}
After the first annotation round, we extended and changed the templates based on the results. 
As mentioned before, we used \textsc{MissScrEv} and \textsc{MissScrPart} labels to mark verbs and noun phrases instantiating events and participants for which no appropriate labels were available in the templates. Based on the instances with these labels (a total of 941 and 1717 instances, respectively), we extended the guidelines to cover the sufficiently frequent cases. 

In order to include new labels for event and participant types, we tried to estimate the number of instances that would fall under a certain label. 
We added new labels according to the following conditions:
\begin{itemize}
	\item For the participant annotations, we added new labels for types that we expected to appear at least 10 times in total in at least 5 different stories (i.e.\ in approximately 5\% of the stories).
	\item For the event annotations, we chose those new labels for event types that would appear in at least 5 different stories. 
\end{itemize}
In order to avoid too fine a granularity of the templates, all other instances of \textsc{MissScrEv} and \textsc{MissScrPart} were re-labeled with \textsc{ScrEv\_other} and \textsc{ScrPart\_other}. 
We also relabeled participants and events from the first annotation phase with \textsc{ScrEv\_other} and \textsc{ScrPart\_other}, if they did not meet the frequency requirements. The event label \textsc{air\_bathroom} (the event of letting fresh air into the room after the bath), for example, was only used once in the stories, so we relabeled that instance to \textsc{ScrEv\_other}.

Additionally, we looked at the DeScript corpus \cite{Wanzare2016}, which contains manually clustered event paraphrase sets for the 10 scenarios that are also covered by InScript (see Section \ref{sec:descript}). Every such set contains event descriptions that describe a certain event type. We extended our templates with additional labels for these events, if they were not yet part of the template.

\subsection{Special Cases}
\label{sec:specialcases}
\paragraph{\textit{Noun-Noun Compounds.}}
Noun-noun compounds were annotated twice with the same label (whole span plus the head noun), as indicated by Example \ref{ex:NounNoun}. 
This redundant double annotation is motivated by potential processing requirements.

\enumsentence{I get my \textit{(wash (cloth $ _{\textsc{\scriptsize ScrPart\_washing\_tools}} ))$, $_{\textsc{\scriptsize ScrPart\_washing\_tools}} $} and put it under the water.}
\label{ex:NounNoun}

\paragraph{\textit{Support Verb Complements.}}
A special treatment was given to support verb constructions such as \textit{take time}, \textit{get home} or \textit{take a seat} in Example \ref{ex:SuppVComp}. The semantics of the verb itself is highly underspecified in such constructions; the event type is largely dependent on the object NP. As shown in Example \ref{ex:SuppVComp}, we annotate the head verb with the event type described by the whole construction
 and label its object with \textsc{SuppVComp} (support verb complement), indicating that it does not have a proper reference.

\enumsentence  {I step into the tub and \textit{take}$ _{\textsc{\scriptsize ScrEv\_sink\_water}} $ a \textit{seat}$ _{\textsc{\scriptsize SuppVComp}} $.} 
\label{ex:SuppVComp}

\paragraph{\textit{Head of Partitive.}}
We used the \textsc{Head\_of\_Partitive} label for the heads in partitive constructions, assuming that the only referential part of the construction is the complement. This is not completely correct, since different partitive heads vary in their degree of concreteness (cf.\ Examples \ref{ex:Partitive1} and \ref{ex:Partitive2}), but we did not see a way to make the distinction sufficiently transparent to the annotators.  

\enumsentence  {Our seats were at the \textit{back}$ _{\textsc{\scriptsize Head\_of\_Partitive}} $ of the train$ _{\textsc{\scriptsize ScrPart\_train}} $.}  
\label{ex:Partitive1}
\enumsentence  {In the library you can always find a \textit{couple}$ _{\textsc{\scriptsize Head\_of\_Partitive}} $ of interesting books$ _{\textsc{\scriptsize ScrPart\_book}} $.}  
\label{ex:Partitive2}

\paragraph{\textit{Mixed Participant Types.}}

Group denoting NPs sometimes refer to groups whose members are instances of different participant types. In Example \ref{ex:MixedCoref1}, the first-person plural pronoun refers to the group consisting of the passenger (\textit{I}) and a non-participant (\textit{my friend}). To avoid a proliferation of event type labels, we labeled these cases with \textsc{Unclear}. 

\enumsentence{\textit{I}$ _{\textsc{\scriptsize {ScrPart\_passenger}}}$ wanted to visit \textit{my}$_{\textsc{\scriptsize{ScrPart\_passenger}}}$ \textit{friend}$ _{\textsc{\scriptsize {NPart}}}$ in New York. ... \textit{We}$_{\textsc{\scriptsize Unclear}}$ met at the train station.} 
\label{ex:MixedCoref1}

We made an exception for the \textsc{Getting a Haircut} scenario, where the mixed participant group consisting of the hairdresser and the customer occurs very often, as in Example \ref{ex:MixedCoref1}. Here, we introduced the additional ad-hoc participant label \textsc{Scr\_Part\_hairdresser\_customer}.

\enumsentence  {While \textit{Susan}$_{\textsc{\scriptsize {ScrPart\_hairdresser}}}$ is cutting \textit{my}$_{\textsc{\scriptsize {ScrPart\_customer}}}$ hair \textit{we}$_{\textsc{\scriptsize Scr\_Part\_hairdresser\_customer}}$ usually talk a bit.}    
\label{ex:MixedCoref2}

\section{Data Analysis}
\subsection{Inter-Annotator Agreement}
\label{sec:iaa}
In order to calculate inter-annotator agreement, a total of 30 stories from 6 scenarios were randomly chosen for parallel annotation by all 4 annotators after the first annotation phase\footnote{We did not test for inter-annotator agreement after the second phase, since we did not expect the agreement to change drastically due to the only slight changes in the annotation schema.}. We checked the agreement on these data using Fleiss' Kappa \cite{fleiss1971measuring}. The results are shown in Figure \ref{table:fleiss_kappa} and indicate moderate to substantial agreement \cite{Landis1977}. Interestingly, if we calculated the Kappa only on the subset of cases that were annotated with script-specific event and participant labels by all annotators, results were better than those of the evaluation on all labeled instances (including also unrelated and related non-script events). This indicates one of the challenges of the annotation task: In many cases it is difficult to decide whether a particular event should be considered a central script event, or an event loosely related or unrelated to the script. 

For coreference chain annotation, we calculated the percentage of pairs which were annotated by at least 3 annotators (qualified majority vote) compared to the set of those pairs annotated by at least one person (see Figure \ref{table:coref_stat}). We take the result of 90.5\% between annotators to be a good agreement.

\subsection{Annotated Corpus Statistics}
\begin{figure}[h]
\centering
\small
\begin{tabular}{|c|c|c|}
\hline
Scenario & Events & Participants \\
\hline
\textsc{bath}     & 20     & 18           \\\hline
\textsc{bicycle}  & 16     & 16           \\\hline
\textsc{bus}      & 17     & 17           \\\hline
\textsc{cake}     & 19     & 17           \\\hline
\textsc{flight}  & 29     & 26           \\\hline
\textsc{grocery}  & 19     & 18           \\\hline
\textsc{haircut}  & 26     & 24           \\\hline
\textsc{library}  & 17     & 18           \\\hline
\textsc{train}    & 15     & 20           \\\hline
\textsc{tree}     & 14     & 15          \\
\hline
\textit{Average} & \textit{19.2} & \textit{18.9} \\
\hline
\end{tabular}
\caption{The number of participants and events in the templates.}
\label{table:template_stats}
\end{figure}

Figure \ref{table:template_stats} gives an overview of the number of event and participant types provided in the templates. \textsc{Taking a flight} and \textsc{getting a haircut} stand out with a large number of both event and participant types, which is due to the inherent complexity of the scenarios. In contrast, \textsc{planting a tree} and \textsc{going on a train} contain the fewest labels. There are 19 event and participant types on average. 

\begin{figure}[h]
\centering
\small
\begin{tabular}{|l|c|c|c|}
\hline
 & avg  & min &  max   \\
 \hline
event annotations in a story       & 15.9 & 1   & 52  \\\hline
event types in a story             & 10.1 & 1   & 23  \\\hline
participant annotations in a story & 52.3 & 16  & 164 \\\hline
participant types in a story       & 10.9 & 2   & 25  \\\hline
coref chains                        & 7.3  & 0   & 23  \\\hline
tokens per chain                   & 6    & 2   & 52 \\
\hline
\end{tabular}
\caption{Annotation statistics over all scenarios.}
\label{table:anno_stats}
\end{figure}

Figure \ref{table:anno_stats} presents overview statistics about the usage of event labels, participant labels and coreference chain annotations. As can be seen, there are usually many more mentions of participants than events. 
For coreference chains, there are some chains that are really long (which also results in a large scenario-wise standard deviation). Usually, these chains describe the protagonist.

We also found again that the \textsc{flying in an airplane} scenario stands out in terms of participant mentions, event mentions and average number of coreference chains.

\begin{figure}[h]
\includegraphics[width=\columnwidth]{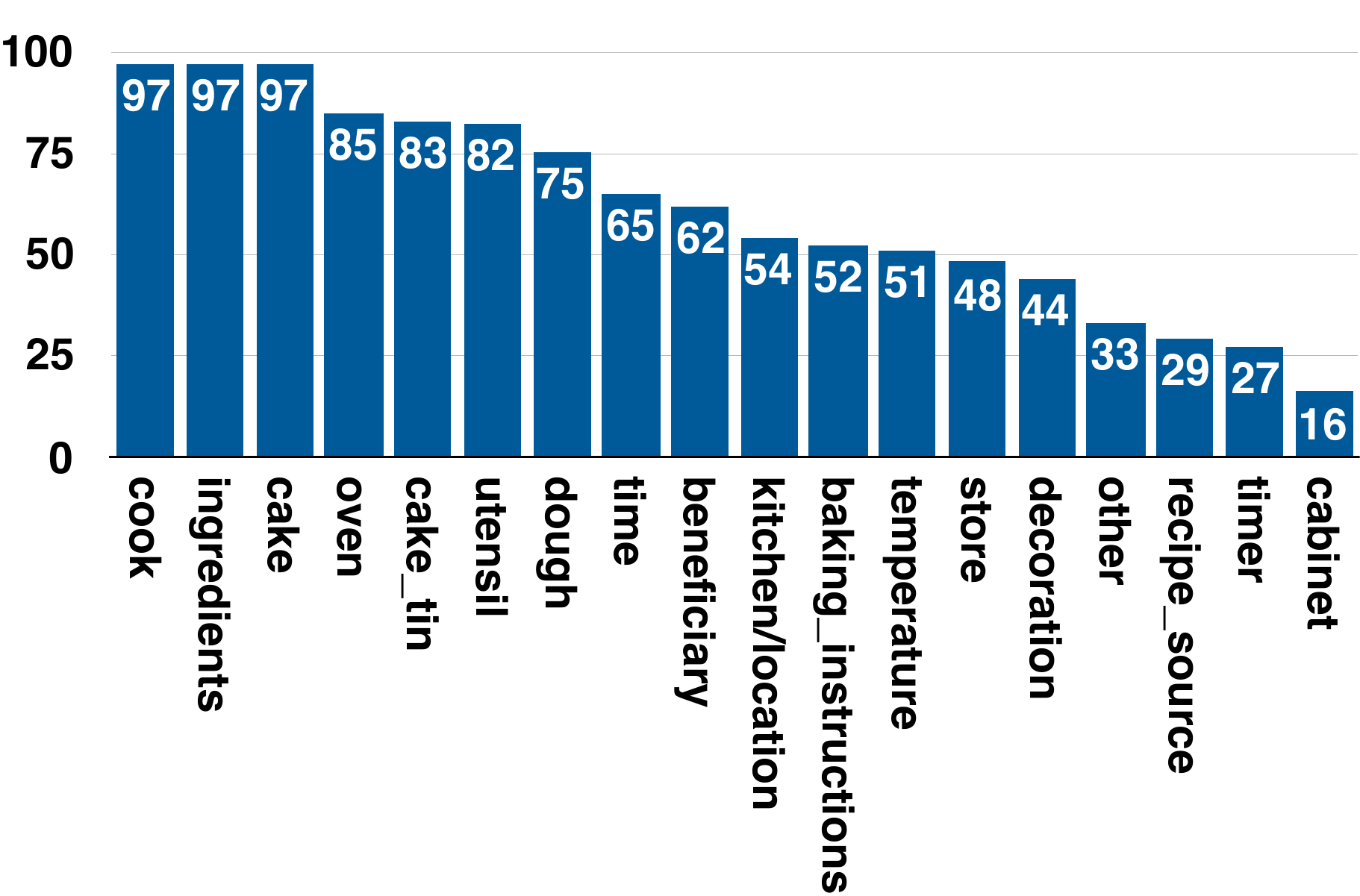}
\caption{The number of stories in the \textsc{baking a cake} scenario that contain a certain participant label.}
\label{fig:evPartDistirbution}
\end{figure}

Figure \ref{fig:evPartDistirbution} shows for every participant label in the \textsc{baking a cake} scenario the number of stories which they occurred in. This indicates how relevant a participant is for the script. As can be seen, a small number of participants are highly prominent: \textsc{cook, ingredients} and \textsc{cake} are mentioned in every story. 
The fact that the protagonist appears most often consistently holds for all other scenarios, where the acting person appears in every story, and is mentioned most frequently.

\begin{figure}[h]
\centering
\begin{minipage}[b]{.39\columnwidth}
\includegraphics[width=1.05\columnwidth]{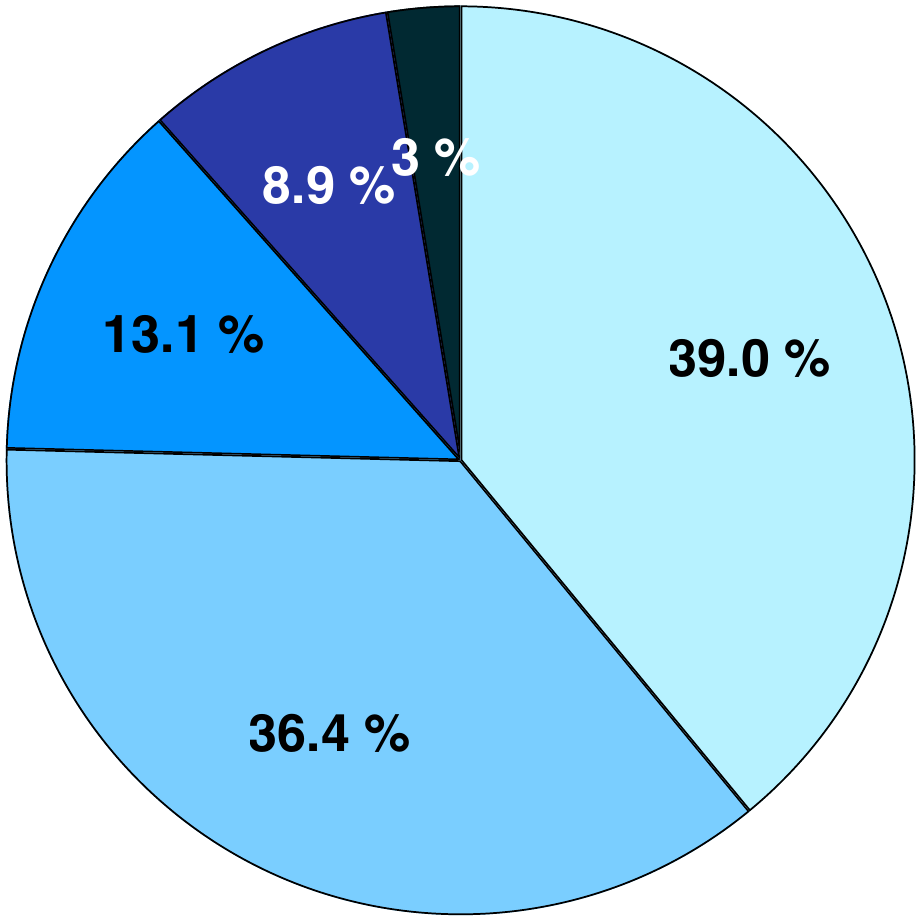}
\end{minipage}
\hspace{0.5cm}
\begin{minipage}[b]{.41\columnwidth}
\includegraphics[width=\columnwidth]{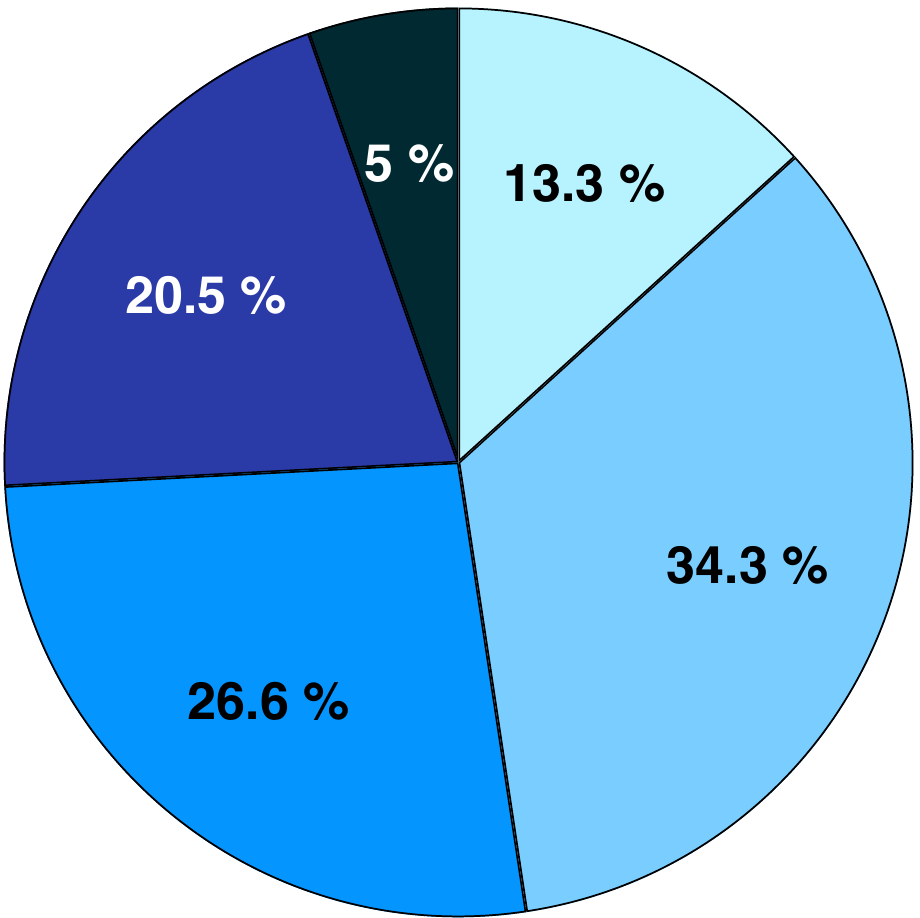}
\end{minipage}
\includegraphics[width=\columnwidth]{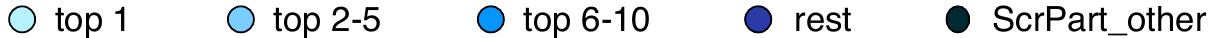}
\caption{Distribution of participants (left) and events (right) for the 1, the top 2-5, top 6-10 most frequently appearing events/participants, \textsc{ScrEv/ScrPart\_Other} and the rest.}
\label{fig:dist}
\end{figure}

Figure \ref{fig:dist} shows the distribution of participant/event type labels over all appearances over all scenarios on average. 
The groups stand for the most frequently appearing label, the top 2 to 5 labels in terms of frequency and the top 6 to 10. \textsc{ScrEv\_other} and \textsc{ScrPart\_other} are shown separately. As can be seen, the most frequently used participant label (the protagonist) makes up about 40\% of overall participant instances. 
The four labels that follow the protagonist in terms of frequency together appear in 37\% of the cases. 
More than 2 out of 3 participants in total belong to one of only 5 labels.

In contrast, the distribution for events is more balanced. 14\% of all event instances have the most prominent event type. 
\textsc{ScrEv\_other} and \textsc{ScrPart\_other} both appear as labels in at most 5\% of all event and participant instantiations: 
The specific event and participant type labels in our templates cover by far most of the instances.

\begin{figure}[h]
\centering
\includegraphics[width=\columnwidth]{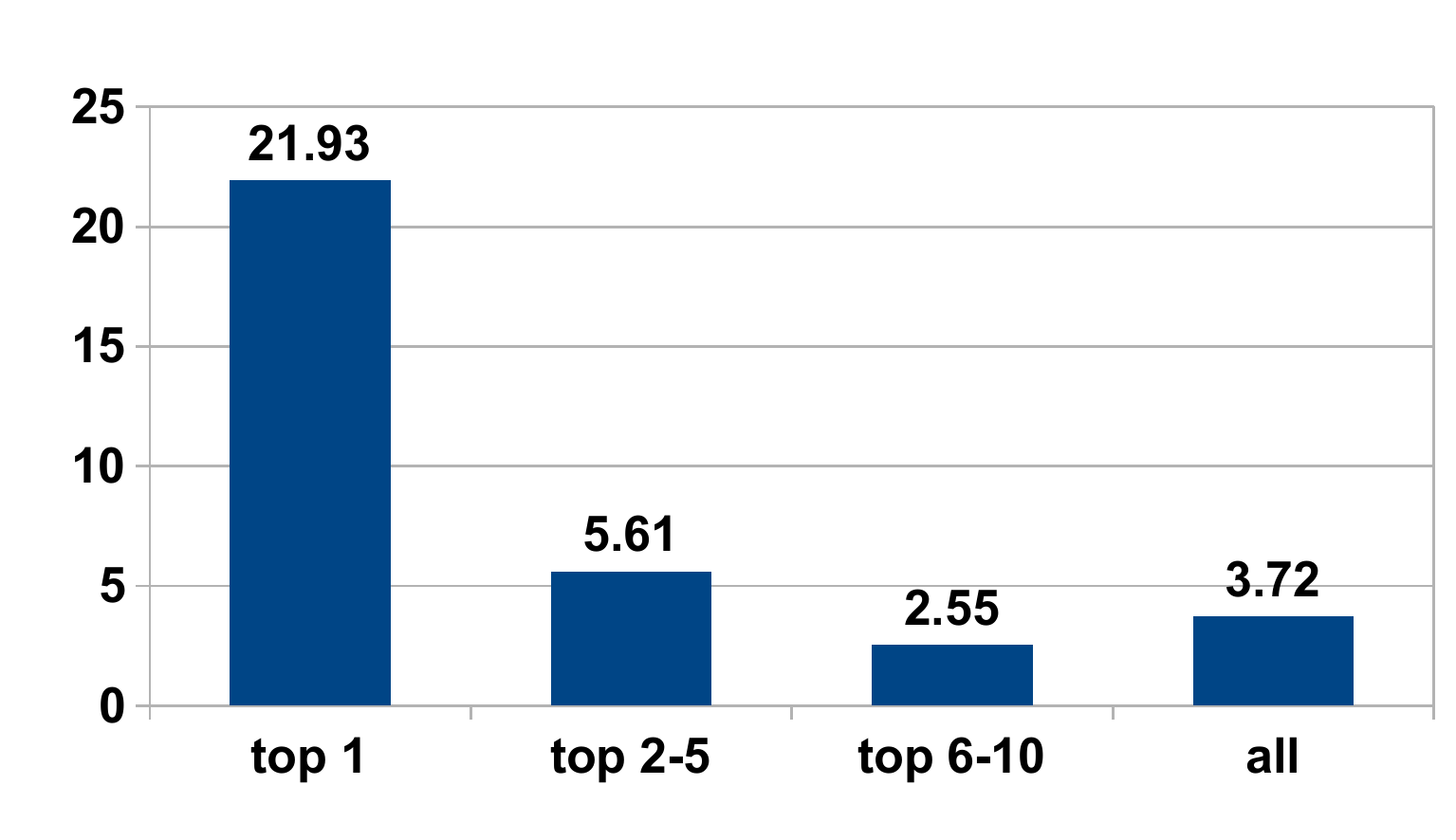}
\caption{Average number of participant mentions for a story, for the first, the top 2-5, top 6-10 most frequently appearing events/participants, and the overall average.}
\label{fig:topn}
\end{figure}

In Figure \ref{fig:topn}, we grouped participants similarly into the first, the top 2-5 and top 6-10 most frequently appearing participant types. The figure shows for each of these groups the average frequency per story, and in the rightmost column the overall average. The results correspond to the findings from the last paragraph.

\subsection{Comparison to the DeScript Corpus}
\label{sec:descript}
As mentioned previously, the InScript corpus is part of a larger research project, in which also a corpus of a different kind, the DeScript corpus, was created. DeScript covers 40 scenarios, and also contains the 10 scenarios from InScript. This corpus contains texts that describe scripts on an abstract and generic level, while InScript contains instantiations of scripts in narrative texts. Script events in DeScript are described in a very simple, telegram-style language 
 (see Figure \ref{fig:scr_ex}). Since one of the long-term goals of the project is to align the InScript texts with the script structure given from DeScript, it is interesting to compare both resources.

The InScript corpus exhibits much more lexical variation than DeScript. Many approaches use the \textit{type-token ratio} to measure this variance. However, this measure is known to be sensitive to text length (see e.g.\ \newcite{Tweedie1998}), which would result in very small values for InScript and relatively large ones for DeScript, given the large average difference of text lengths between the corpora. Instead, we decided to use the \textit{Measure of Textual Lexical Diversity (MTLD)} (\newcite{McCarthy2010}, \newcite{McCarthy2005}), which is familiar in corpus linguistics. This metric measures the average number of tokens in a text that are needed to retain a type-token ratio above a certain threshold. If the \textit{MTLD} for a text is high, many tokens are needed to lower the type-token ratio under the threshold, so the text is lexically diverse. In contrast, a low \textit{MTLD} indicates that only a few words are needed to make the type-token ratio drop, so the lexical diversity is smaller. We use the threshold of 0.71, which is proposed by the authors as a well-proven value.

\begin{figure}[h]
\centering
\includegraphics[width=\columnwidth]{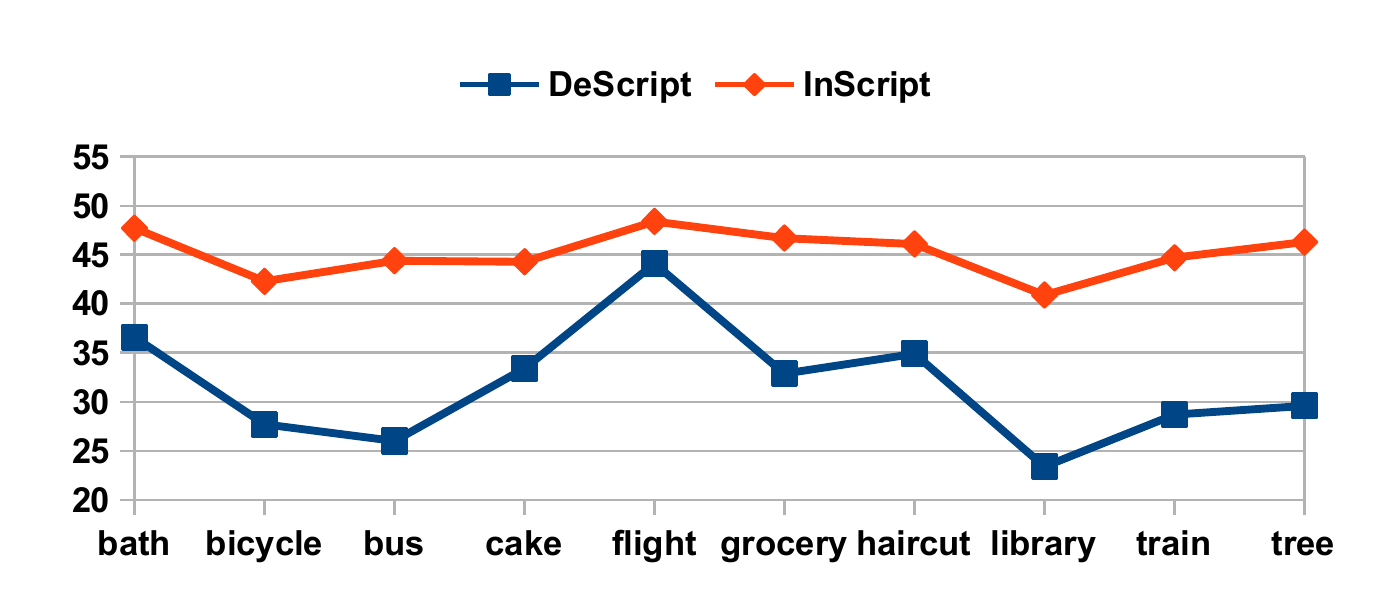}
\caption{MTLD values for DeScript and InScript, per scenario.}
\label{fig:mtld}
\end{figure}

Figure \ref{fig:mtld} compares the lexical diversity of both resources. As can be seen, the InScript corpus with its narrative texts is generally much more diverse than the DeScript corpus with its short event descriptions, across all scenarios. For both resources, the \textsc{flying in an airplane} scenario is most diverse (as was also indicated above by the mean word type overlap). However, the difference in the variation of lexical variance of scenarios is larger for DeScript than for InScript. Thus, the properties of a scenario apparently influence the lexical variance of the event descriptions more than the variance of the narrative texts. 

We used entropy \cite{Shannon1948} over lemmas 
to measure the variance of lexical realizations for events. 
We excluded events for which there were less than 10 occurrences in DeScript or InScript. Since there is only an event annotation for 50 ESDs per scenario in DeScript, we randomly sampled 50 texts from InScript for computing the entropy to make the numbers more comparable.
\begin{figure}[h]
\centering
\includegraphics[width=\columnwidth]{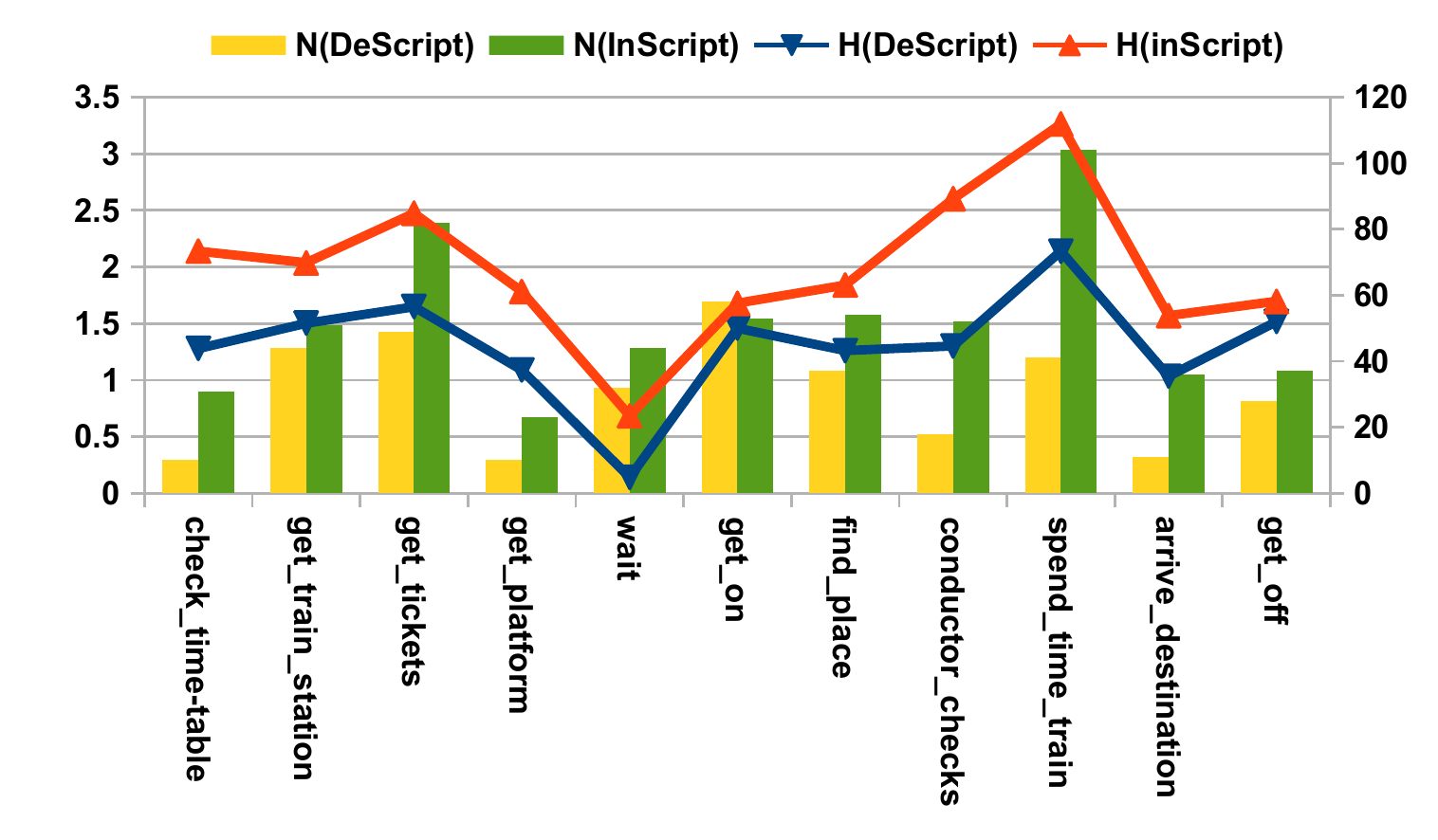}
\caption{Entropy over verb lemmas for events (left y-axis, H(x)) in the \textsc{going on a train scenario}. Bars in the background indicate the absolute number of occurrence of instances (right y-axis, N(x)).}
\label{fig:entropy}
\end{figure}

Figure \ref{fig:entropy} shows as an example the entropy values for the event types in the \textsc{going on a train} scenario. As can be seen in the graph, the entropy for InScript is in general higher than for DeScript. In the stories, a wider variety of verbs is used to describe events. There are also large differences between events: While \textsc{wait} has a really low entropy, \textsc{spend\_time\_train} has an extremely high entropy value. This event type covers many different activities such as reading, sleeping etc.

\section{Conclusion}
In this paper we described the InScript corpus of 1,000 narrative texts annotated with script structure and coreference information. We described the annotation process, various difficulties encountered during annotation and different remedies that were taken to overcome these. One of the future research goals of our project is also concerned with finding automatic methods for text-to-script mapping, i.e.\ for the alignment of text segments with script states. We consider InScript and DeScript together as a resource for studying this alignment. The corpus shows rich lexical variation and will serve as a unique resource for the study of the role of script knowledge in natural language processing.  

\section*{Acknowledgements}
This research was funded by the German Research Foundation (DFG) as part of SFB 1102 'Information Density and Linguistic Encoding'.

%
%
%
%
%

\section{References}
\bibliographystyle{lrec2016}
\bibliography{refrences}

\begin{thebibliography}{}

\bibitem[\protect\citename{Barr and Feigenbaum}1981]{Barr1981}
Barr, A. and Feigenbaum, E.~A.
\newblock (1981).
\newblock {\em The Handbook of Artificial Intelligence}.
\newblock Addison-Wesley.

\bibitem[\protect\citename{Chambers and Jurafsky}2008]{Chambers2008}
Chambers, N. and Jurafsky, D.
\newblock (2008).
\newblock Unsupervised learning of narrative event chains.
\newblock {\em Proceedings of ACL-08}.

\bibitem[\protect\citename{Chambers and Jurafsky}2009]{Chambers2009}
Chambers, N. and Jurafsky, D.
\newblock (2009).
\newblock Unsupervised learning of narrative schemas and their participants.
\newblock {\em Proceedings of the 47th Annual Meeting of the ACL and the 4th
  IJCNLP of the AFNLP}.

\bibitem[\protect\citename{Cullingford}1978]{cullingford1978script}
Cullingford, R.~E.
\newblock (1978).
\newblock Script application: Computer understanding of newspaper stories.
\newblock Technical report, DTIC Document.

\bibitem[\protect\citename{Fleiss}1971]{fleiss1971measuring}
Fleiss, J.~L.
\newblock (1971).
\newblock Measuring nominal scale agreement among many raters.
\newblock {\em Psychological bulletin}, 76(5):378.

\bibitem[\protect\citename{Landis and Koch}1977]{Landis1977}
Landis, J.~R. and Koch, G.~G.
\newblock (1977).
\newblock {The Measurement of Observer Agreement for Categorical Data}.
\newblock {\em {Biometrics}}, 33(1):pp. 159--174.

\bibitem[\protect\citename{McCarthy and Jarvis}2010]{McCarthy2010}
McCarthy, P.~M. and Jarvis, S.
\newblock (2010).
\newblock Mtld, vocd-d, and hd-d: A validation study of sophisticated
  approaches to lexical diversity assessment.
\newblock {\em Behavior Research Methods}.

\bibitem[\protect\citename{McCarthy}2005]{McCarthy2005}
McCarthy, P.~M.
\newblock (2005).
\newblock {\em An Assessment of the Range and Usefulness of Lexical Diversity
  Measures and the Potential of the Measure of Textual, Lexical Diversity
  (MTLD)}.
\newblock {Ph.D.} thesis, The University of Memphis.

\bibitem[\protect\citename{Miikkulainen}1995]{miikkulainen1995script}
Miikkulainen, R.
\newblock (1995).
\newblock Script-based inference and memory retrieval in subsymbolic story
  processing.
\newblock {\em Applied Intelligence}, 5(2):137--163.

\bibitem[\protect\citename{Modi and Titov}2014]{modi2014inducing}
Modi, A. and Titov, I.
\newblock (2014).
\newblock Inducing neural models of script knowledge.
\newblock In {\em CoNLL}, volume~14, pages 49--57.

\bibitem[\protect\citename{Mueller}2004]{mueller2004understanding}
Mueller, E.~T.
\newblock (2004).
\newblock Understanding script-based stories using commonsense reasoning.
\newblock {\em Cognitive Systems Research}, 5(4):307--340.

\bibitem[\protect\citename{Raisig \bgroup et al.\egroup }2009]{VanDerMeer2009}
Raisig, S., Welke, T., Hagendorf, H., and Van Der~Meer, E.
\newblock (2009).
\newblock Insights into knowledge representation: The influence of amodal and
  perceptual variables on event knowledge retrieval from memory.
\newblock {\em Cognitive Science}, 33(7):1252--1266.

\bibitem[\protect\citename{Regneri \bgroup et al.\egroup }2010]{Regneri:2010}
Regneri, M., Koller, A., and Pinkal, M.
\newblock (2010).
\newblock Learning script knowledge with web experiments.
\newblock In {\em Proceedings of the 48th Annual Meeting of the Association for
  Computational Linguistics}, ACL '10, pages 979--988, Stroudsburg, PA, USA.
  Association for Computational Linguistics.

\bibitem[\protect\citename{Regneri}2013]{Regneri2013}
Regneri, M.
\newblock (2013).
\newblock {\em Event Structures in Knowledge, Pictures and Text}.
\newblock {Ph.D.} thesis, Universit\"at des Saarlandes.

\bibitem[\protect\citename{Rudinger \bgroup et al.\egroup
  }2015]{rudinger2015learning}
Rudinger, R., Demberg, V., Modi, A., Van~Durme, B., and Pinkal, M.
\newblock (2015).
\newblock Learning to predict script events from domain-specific text.
\newblock {\em Lexical and Computational Semantics (* SEM 2015)}, page 205.

\bibitem[\protect\citename{Shannon}1948]{Shannon1948}
Shannon, C.~E.
\newblock (1948).
\newblock {A} {M}athematical {T}heory of {C}ommunication.
\newblock {\em The Bell System Technical Journal}, 27(3):379--423.

\bibitem[\protect\citename{Singh \bgroup et al.\egroup }2002]{singh2002open}
Singh, P., Lin, T., Mueller, E.~T., Lim, G., Perkins, T., and Zhu, W.~L.
\newblock (2002).
\newblock Open mind common sense: Knowledge acquisition from the general
  public.
\newblock In {\em On the move to meaningful internet systems 2002: CoopIS, DOA,
  and ODBASE}, pages 1223--1237. Springer.

\bibitem[\protect\citename{Tweedie and Baayen}1998]{Tweedie1998}
Tweedie, F.~J. and Baayen, R.~H.
\newblock (1998).
\newblock {How Variable May a Constant Be? Measures of Lexical Richness in
  Perspective}.
\newblock {\em Computers and the Humanities}, 32(5):323--352.

\bibitem[\protect\citename{Wanzare \bgroup et al.\egroup }2016]{Wanzare2016}
Wanzare, L. D.~A., Zarcone, A., Thater, S., and Pinkal, M.
\newblock (2016).
\newblock A crowdsourced database of event sequence descriptions for the
  acquisition of high-quality script knowledge.
\newblock {\em Proceedings of the Tenth International Conference on Language
  Resources and Evaluation (LREC'16)}.

\bibitem[\protect\citename{Yimam \bgroup et al.\egroup }2013]{YimamGCB13}
Yimam, S.~M., Gurevych, I., de~Castilho, R.~E., and Biemann, C.
\newblock (2013).
\newblock {WebAnno: A Flexible, Web-based and Visually Supported System for
  Distributed Annotations}.
\newblock In {\em ACL (Conference System Demonstrations)}, pages 1--6.

\end{thebibliography}
\end{document}